\documentclass[runningheads]{llncs}

\usepackage{eccv}

\usepackage{eccvabbrv}

\usepackage{graphicx}
\usepackage{booktabs}

\usepackage[accsupp]{axessibility}  

\usepackage[pagebackref,breaklinks,colorlinks,citecolor=eccvblue]{hyperref}

\usepackage{orcidlink}

\usepackage{bm}
\usepackage{bbm}

\usepackage{soul}

\usepackage{graphicx}
\usepackage{tabularx}
\usepackage{subcaption}
\usepackage{multirow}
\usepackage{mathrsfs}
\usepackage{amsmath}

\begin{document}

\title{BAGS: Building  Animatable  Gaussian Splatting from a Monocular Video with Diffusion Priors} 

\titlerunning{BAGS}

\author{
Tingyang Zhang\thanks{Joint first authors with equal contributions to this work.} \inst{1},
Qingzhe Gao\textsuperscript{$\star$}\inst{2,1},
Weiyu Li\inst{3},
Libin Liu\inst{1},
Baoquan Chen\inst{1},
}

\authorrunning{Zhang et al.}

\institute{
National Key Lab of General AI, Peking University, China 
\and
Shandong University, China 
\and
The Hong Kong University of Science and Technology
\\
\url{https://talegqz.github.io/BAGS/} 
\\
\email{2100012962@stu.pku.edu.cn,
\{gaoqingzhe97,weiyuli.cn\}@gmail.com,
\{libin.liu, baoquan\}@pku.edu.cn}
}


\maketitle

\begin{figure}[h!]
  \centering
  \includegraphics[width=0.95\linewidth]{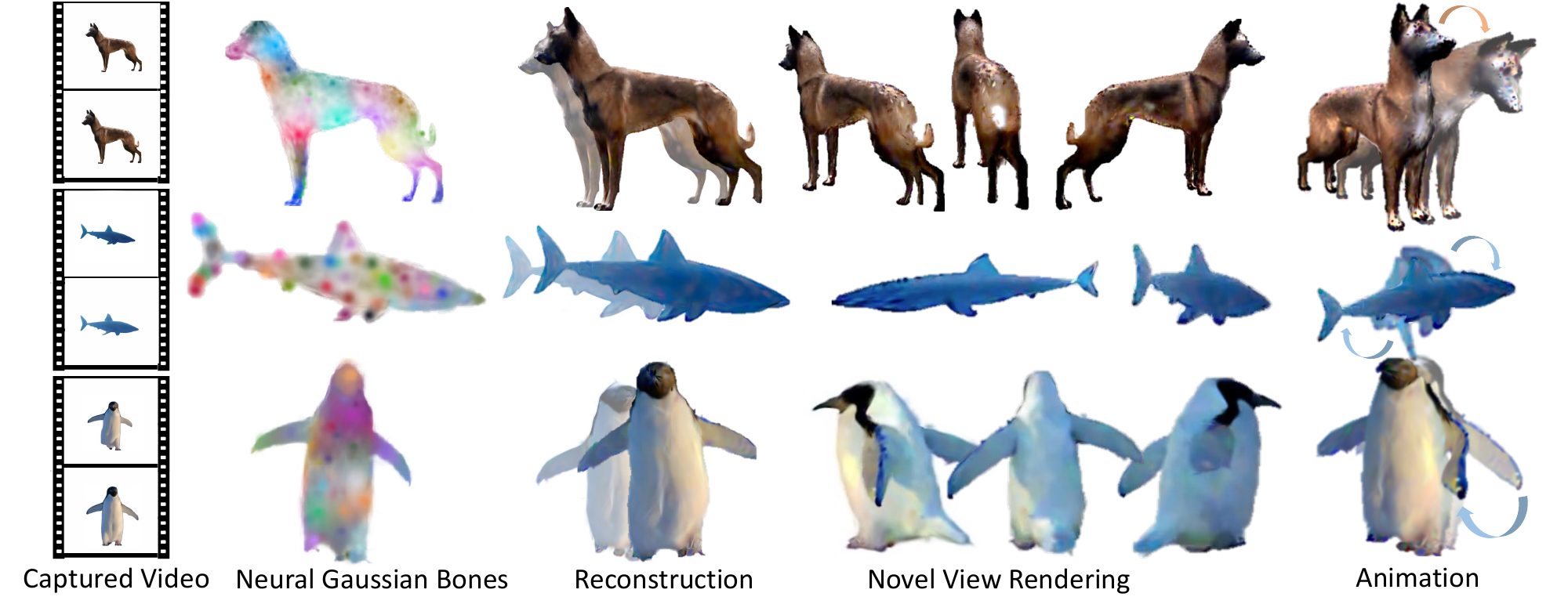}
  \caption{
  Given a single casual video, our method constructs an animatable 3D Gaussian Splatting model with diffusion priors. This not only compensates for unseen view information but also enables fast training and real-time rendering.
  }
  \label{fig:teaser}
\end{figure}

\begin{abstract}
 Animatable  3D reconstruction has significant applications across various fields, primarily relying on artists' handcraft creation. Recently, some studies have successfully constructed animatable 3D models from monocular videos. However, these approaches require sufficient view coverage of the object within the input video
 and typically necessitate significant time and computational costs for training and rendering. This limitation restricts the practical applications. 
 In this work, we propose a method to build animatable 3D Gaussian Splatting from monocular video with diffusion priors. 
 The 3D Gaussian representations significantly accelerate the training and rendering process, and the diffusion priors allow the method to learn 3D models with limited viewpoints.
 We also present the rigid regularization to enhance the utilization of the priors.
 We perform an extensive evaluation across various real-world videos, demonstrating its superior performance compared to the current state-of-the-art methods.
  \keywords{Gaussian Splatting \and Animation \and Diffusion models}
\end{abstract}

\section{Introduction}
Animatable 3D models play an important role in diverse fields such as augmented and virtual reality (AR/VR), gaming, and digital content creation, offering advanced motion control and flexibility for various applications.
However, the process of acquiring these models poses significant challenges. Traditional methods are not only time-consuming and costly but also heavily reliant on the expertise of artists, which can result in models that often fall short of achieving realistic accuracy.

An alternative approach involves directly reconstructing articulated objects from real-world data. By employing template-based parametric models~\cite{base_SMAL, SMPL_X, base_Flame, base_SMPL}, some works\cite{ tb_app_lions_tigers_bears, tb_app_bird} have successfully reconstructed animatable models from videos, even with a limited number of images. However, these parametric models depend on extensive registered 3D scans of humans or animals. This dependence presents great challenges when 3D scan data is unavailable.

Alternatively, videos can provide motion information, facilitating the learning of some aspects of motion and structure~\cite{motion_co_part, motion_gao2021unsupervised}. Utilizing videos, some studies~\cite{tf_video_new_dovewu2023dove, video_new_li2022tava, Banmo_yang2022banmo,tf_kokkinos2021point,video_new_tan2023distilling} learned to reconstruct animatable objects. 
Among these approaches, some works still depend on category-level information~\cite{video_new_tan2023distilling} to aid in reconstruction,  such as key points~\cite{video_new_li2022tava}.
Recently,
BANMo~\cite{Banmo_yang2022banmo} achieved state-of-the-art results in reconstructing non-rigid, animatable 3D models from single-object videos without relying on template shape priors and category-level information. However, BANMo suffers from slow training and rendering time due to its use of NeRF~\cite{NeRF_mildenhall2021nerf}. Moreover, an individual video may not always contain sufficient information to accurately reconstruct a given object. Consequently, BANMo incorporates several videos of the same objects, such as filming a family member or a pet over several months or years. Nonetheless, access to such extensive video collections may not be feasible for all objects.

To address these challenges, we introduce \textbf{BAGS}, designed to \textbf{B}uild \textbf{A}nima-table 3D \textbf{G}aussian \textbf{S}platting with diffusion priors.
We propose animatable Gaussian Splitting driven by neural bone, enabling rapid training and rendering.
To address the challenge posed by insufficient view coverage, we incorporate the diffusion priors as supervision for our model. 
However, a na\"ive application of the priors may lead to the emergence of artifacts, as the diffusion model may not guarantee accuracy and consistency across all time.
Consequently, we introduce a rigid regularization technique aimed at optimizing the utilization of the priors.

In experiments, we demonstrate that our approach surpasses the state-of-the-art methods in terms of geometry, appearance, and animation quality with in-the-wild videos.
In summary, our contributions are:
\begin{itemize}
\item We present BAGS, a framework for constructing animatable 3D Gaussian Splatting incorporating diffusion priors, which achieves state-of-the-art performance, alongside rapid training and real-time rendering.
\item We integrate diffusion priors to compensate for the unseen view information in casual videos. Furthermore, we introduce a rigid regularization technique to enhance the utilization of the priors.
\item  We collect a dataset from in-the-wild videos to evaluate our method. Both qualitative and quantitative results demonstrate that BAGS achieves superior performance compared to baseline models.
\end{itemize}

\section{Related work}
\subsection{Animatable model reconstruction}
As parametric models~\cite{base_Flame,base_SMAL, SMPL_X,base_SMPL} develop, some research~\cite{tb_app_bird,tb_app_dogs,tb_app_face,tb_app_gao2023neural,tb_app_gart,tb_app_lions_tigers_bears,tb_app_rueegg2022barc,tb_app_zuffi2019three} can recover 3D shapes and motions for the human body, face, and animals, even with a few or a single image as input. However, these parametric models are built from large volumes of ground-truth 3D data. It is challenging to apply them to arbitrary categories where  3D data of scan is unavailable.
Alternatively, videos can provide temporal information that aids in understanding motion~\cite{motion_co_part, motion_gao2021unsupervised}, and some research~\cite{video_old_bregler2000recovering,video_old_goel2020shape,video_old_gotardo2011non,video_old_kong2019deep,video_old_kumar2020non,video_old_sand2008particle,video_old_sidhu2020neural,video_old_yang2021lasr,video_new_viser_yang2021viser} relies on video to recover shape and motion. Nevertheless, these approaches may still result in blurry geometry and unrealistic articulations. Inspired by the promising outcomes of dynamic neural representations~\cite{nerf_barf_in2021barf,nerf_d_gao2021dynamic,nerf_d_liu2023robust,nerf_park2021nerfies, NeRF_mildenhall2021nerf, D_neRF_pumarola2021d}, some research~\cite{tf_video_new_dovewu2023dove, Banmo_yang2022banmo,video_new_ppr_yang2023ppr,video_new_li2022tava,video_new_ppr_yang2023ppr,video_new_tan2023distilling,video_new_viser_yang2021viser} has demonstrated the ability to reconstruct appearances with enhanced quality. However, these methods require the input video to cover all views of the object comprehensively. Among these, Banmo is the most relevant to our method. In addition to the above problem, BANMo~\cite{Banmo_yang2022banmo} relies on NeRF~\cite{NeRF_mildenhall2021nerf}, which results in expensive training and rendering times.
Our method, in contrast, leverages  3D Gaussian Splatting~\cite{3dgs_kerbl20233d} based on video input, enabling quick training and real-time rendering. Additionally, we employ a diffusion model to address the lack of viewpoint information.

\subsection{Reconstruction with Priors}
Leveraging category-level priors,
several methods~\cite{tf_app_kanazawa2018learning,tf_app_tulsiani2020implicit,tf_artic_3dyao2024artic3d,tf_kokkinos2021point,tf_yao2022lassie,video_old_goel2020shape} have been proposed for creating animatable 3D models without the use of templates, by reconstructing shapes and poses from 2D image datasets. These approaches employ weak supervision techniques, including key points~\cite{video_new_li2022tava} and object silhouettes, obtained from offthe-shelf models or through human annotations. However, these approaches still necessitate category-level information and exhibit limitations in scenarios with sparse data inputs.

Diffusion models~\cite{diff_ramesh2021zero,diff_rombach2022high,diff_saharia2205photorealistic,diff_song2020score,diff_zhang2023adding} have demonstrated remarkable capabilities in generating realistic images beyond category level. Recent studies~\cite{diff_3d_raj2023dreambooth3d,diff3d_lin2023magic3d,diff3d_metzer2023latent,diff3d_poole2022dreamfusion,diff3d_ren2023dreamgaussian4d,diff3d_richardson2023texture,diff3d_singer2023text,diff3d_wang2023score,zero123_liu2023zero} have explored utilizing 2D diffusion models to create 3D assets from text prompts or input images. Moreover, some methods~\cite{diffrecon_wu2023reconfusion,diffrecon_wynn2023diffusionerf,diffrecon_zhou2023sparsefusion,zero123_liu2023zero} reconstruct static scenes from sparse image inputs. Our method not only employs diffusion priors but also integrates motion information from videos to reconstruct animatable models.

\paragraph{Concurrent work} DreaMo~\cite{DreaMo_tu2023dreamo} similarly utilizes diffusion priors and video information to reconstruct animatable models. Unlike DreaMo, which employs NeRF~\cite{NeRF_mildenhall2021nerf}, our method incorporates 3DGS\cite{3dgs_kerbl20233d}, achieving superior training efficiency and rendering speed. Additionally, while DreaMo primarily uses diffusion for geometric supervision, our approach uses it to enhance appearance in uncovered viewpoints.

\section{Method}
\begin{figure}[h!]
  \centering
  \includegraphics[width=0.95\linewidth]{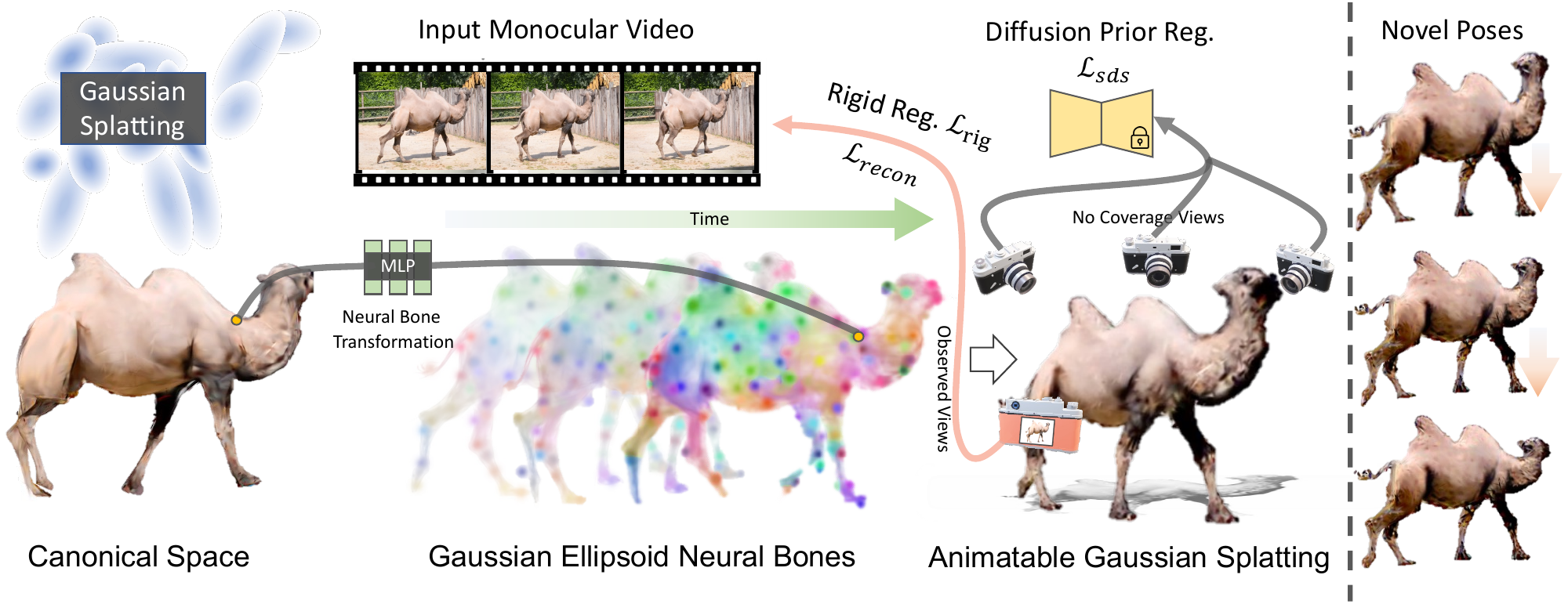}
  \caption{
  Overview of our method. 
  We construct a canonical space using Gaussian Splatting. 
  In the absence of a templated parametric model, we develop a neural bones representation to animate the canonical space to match the input video. Additionally, we utilize a diffusion model to address unseen view information and apply a rigid constraint to facilitate training. After training, the model can be manually manipulated to achieve novel pose rendering.
  }
  \label{fig:pipeline}
\end{figure}

Given a single video of a moving object, our method aims to construct an animatable 3D model of this object. 
We outline the necessary preliminaries in Section~\ref{sec:Preliminaries}. 
Subsequently, we introduce our novel neural bone and LBS structure, detailing its utilization for animating the Gaussian Splatting in Section~\ref{sec:neural_bone}. 
Finally, we describe the diffusion priors to compensate for the unseen views of the object and apply additional regularization techniques to improve the training procedure in Section~\ref{sec:diff}. The pipeline of our approach is illustrated in Figure~\ref{fig:pipeline}.
\subsection{Preliminaries}
\label{sec:Preliminaries}
\subsubsection{Gaussian Splatting}
3D Gaussian Splatting~\cite{3dgs_kerbl20233d} is a scene representation that allows high-quality real-time rendering. The representation is parameterized by a set of static 3D Gaussians. The $i^{th}$ Gaussian, denoted as $g_i = \{t_i, R_i, S_i, o_i, c_i\}$, is defined by a translation $t_i \in \mathbb{R}^3$, a rotation  $R_i \in SO(3)$, a per-axis scale, represented by a diagonal matrix $S_i \in \mathbb{R}^3$, an opacity $o_i \in \mathbb{R}$, and a color $c_i \in \mathbb{R}^3$. The spatial extent of each Gaussian is defined in 3D by its covariance matrix $\Sigma_i = R_i S_i S_i^T R_i^T.$

For rendering these 3D Gaussians, the method~\cite{diff3d_ren2023dreamgaussian4d} first transforms them from the world to the camera space and then projects them onto a 2D screen plane. This projection employs the covariance matrix $\Sigma^{2D}_i = JW\Sigma_iW^TJ^T$, where $J \in \mathbb{R}^{2\times3}$ denotes the Jacobian of an affine approximation of the projective transformation, and $W \in \mathbb{R}^{3\times3}$ is the viewing transformation matrix. The color $C_{\mathbf{p}}$ of a pixel $\mathbf{p}$ is determined through alpha blending of the $\mathbf{N}$ ordered Gaussians contributing to that pixel, as:
\begin{equation}
C(\mathbf{p})= \sum_{i \in \mathbf{N}} c_i \alpha_i \prod_{j=1}^{i-1} (1 - \alpha_j)
\label{eq:gau_rendering}
\end{equation}
The term $\alpha_i$ is derived using the 2D covariance matrix $\Sigma^{2D}_i$ and calculated by the per-Gaussian opacity $o_i$.

\subsubsection{Linear Blender Skinning}
Given a 3D point $\mathbf{x}_c$, we can transform the point $\mathbf{x}_t$ by
the Linear Blend Skinning (LBS) function, which is defined as:
\begin{align}
    \mathbf{x}_t =\text{LBS}(\mathbf{x}_c, \{\mathbf{\Delta B}_b\}^B_{b=1}, \omega_{\mathbf{x_c}})=(\sum_{b=1}^{B}\omega_{\mathbf{x_c}}^{b}\mathbf{\Delta B}_b)\mathbf{x}_c
    \label{eq_LBS}
\end{align}
where $\{\mathbf{\Delta B}_b\}^B_{b=1}$ are the rigid bone transformations,
and $\omega_{\mathbf{x}_c}^b$ represents the skinning weight of point $\mathbf{x}_c$ with respect to bone $\mathbf{B}_b$.

\begin{figure}[h!]
  \centering
  \includegraphics[width=0.95\linewidth]{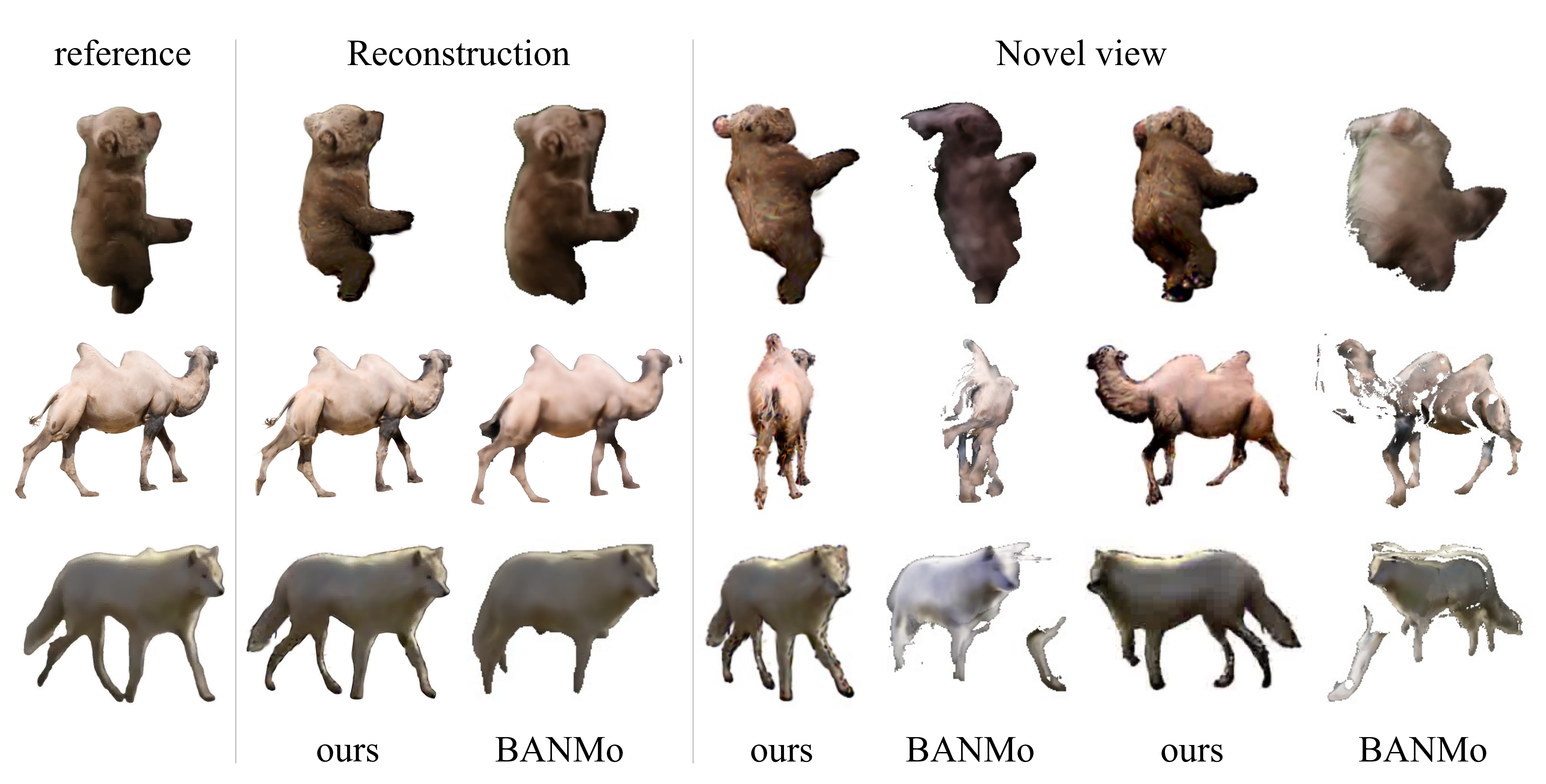}
  \caption{
  Qualitative results.
  Compared with BANMo~\cite{Banmo_yang2022banmo}, our method demonstrates superior fidelity to the input image, exhibiting enhanced geometric detail and texture richness. Moreover, our method achieves better performance in novel view synthesis, where BANMo fails to produce reasonable 3D shapes, instead getting merely a 2D plane that overfits the input video.
  }
  \label{fig:compare}
\end{figure}

\subsection{Neural Bone and Skinning Weight}
\label{sec:neural_bone}
The bone transformation mentioned can be estimated using parametric model parameters, such as SMPL~\cite{base_SMPL} and SMAL~\cite{base_SMAL}.
Building on this, numerous work ~\cite{tb_app_bird,tb_app_dogs,tb_app_gart} successfully constructed animatable 3D models.
The key idea of these methods involves constructing a canonical space and utilizing the parametric model to drive this canonical space to the target pose.
In contrast to these studies, our paper does not depend on such priors knowledge.
Inspired by BANMo~\cite{Banmo_yang2022banmo}, we develop the neural bone and skinning weight approach.


The set of neural bones, denoted by $\{\mathbf{B}_b\}$, is defined using Gaussian ellipsoids, where each bone $\mathbf{B}_b$ is represented by center $C_{\mathbf{B}_b} \in \mathbb{R}^{3\times1}$, diagonal scale matrices $D_{\mathbf{B}_b} \in \mathbb{R}^{3\times3}$ and  rotation $M_{\mathbf{B}_b} \in \mathbb{R}^{3\times3}$. 
In practice, for each time point $t$, the MLPs is employed to predict the bone parameters based on the positional embedding, $\gamma(t)$:
\begin{align}
C_{\mathbf{B}}^t=\text{MLP}_C(\gamma(t)), \quad D_{\mathbf{B}}^t=\text{MLP}_D(\gamma(t)), \quad M_{\mathbf{B}}^t=\text{MLP}_M(\gamma(t)), \quad 
    \label{base_weight}
\end{align}
Without loss of generality, the bone information in canonical space can be determined through a learnable embedding.

The transformation of each bone is determined by its rotation and position, whereas its influence is determined by the scale matrix. 
More specifically, akin to the approach in LASR~\cite{video_old_yang2021lasr},
the weight $W_{\mathbf{x}}^{b}$ of given point $s$
the relationship between a given point $\mathbf{x}$  and a bone $\mathbf{B}_b$ can be calculated using the Mahalanobis distance:
\begin{align}
    W_{\mathbf{x}}^{b} = (x-C_b)^T M_b^T D M_b (x-C_b)
    \label{softmax_weight}
\end{align}
We then apply the softmax function to normalize the weight, resulting in the final skinning weights:
\begin{align}
    \{\omega_{\mathbf{x}}^b\} = \text{softmax}(\{W_{\mathbf{x}}^{b}\})
\end{align}

Thanks to the explicit modeling provided by Gaussian splatting, we can directly apply Linear Blend Skinning (LBS) as described in Equation \eqref{eq_LBS} to drive the model in canonical space to the target pose.
For a given target pose in time $t$, we can readily obtain 
$g_i^t$, which transforms the 3D Gaussian $g_i^c$ in canonical space,as follows:
\begin{equation}
    g_i^t = \text{LBS}(g^c_i, \{\mathbf{\Delta B}_b\}^B_{b=1}, \omega_{g^c_i})  =(\sum_{b=1}^{B}\omega_{g^c_i}^{b}\mathbf{\Delta B}_b)g^c_i = A^t_i g^c_i 
\end{equation}
where the LBS weight $\omega(g_i)$ can be estimated using Equations \eqref{base_weight} and \eqref{softmax_weight}.
Moreover, the bone transformations $\{\mathbf{\Delta B}_b\}^B_{b=1}$ can be derived from the rotations and translations between bones in the canonical space and the target pose.
     

Subsequently, the rotation and scale of $g_i^t$ can be obtained through the changes in the covariance matrix:
\begin{equation}
    \Sigma_i^t = J^t_i\Sigma_i^c ({J^t_i})^{T}, \quad J^t_i = A^t_i + \frac {\partial A^t_i}{\partial g_i^c} g_i^c
\end{equation}
The color and opacity of the Gaussian are assumed to remain constant. Consequently, we can utilize Equation \eqref{eq:gau_rendering} to render the final color $C_{\mathbf{p}_t}$ of pixel $\mathbf{p}_t$ at time $t$.

In summary, given a time $t$ and arbitrary camera parameters $P$, we can render the corresponding image $\mathbf{\Tilde{I}}^t_P $ with our model.
We denote this process by $\mathbf{\Tilde{I}}^t_P = F_{\phi}(P,t)$, where $F$ represents our model, and $\phi$ denotes the model's parameters.

\subsection{Diffusion Priors}
\label{sec:diff}
The primary challenge in constructing an animatable object from a video lies in the insufficiency of the input video to cover the object comprehensively. If the video does not encompass enough perspectives, it lacks the necessary information for an accurate reconstruction of the subject. BANMo~\cite{Banmo_yang2022banmo} addresses this issue by utilizing casually collected videos.
However, collecting suitable videos can be challenging, therefore, we employ the diffusion priors to address this issue.

Conditioned on the input image, we employ ImageDream~\cite{imagedream_wang2023imagedream} to synthesize novel view images and utilize Score Distillation Sampling (SDS) to supervise our model. Given a time $t$, and the input image $\mathbf{I}_{t}$, our model randomly generates multiple views. The SDS loss can be expressed as follows:
\begin{align}
    \nabla_\phi \mathcal{L_\text{sds}} = \mathbb{E}_{t,\tau,\epsilon,P} [(\epsilon_\theta(F_\phi(P,t);\tau,I_{t},P) - \epsilon) \frac{\partial F}{\partial \phi}]
\end{align}
Where the $\epsilon$ represents random noise, $\epsilon_{\theta}$ denotes the noise predicted by the diffusion model, and $P$ is the random camera viewpoint.
$\tau$ 
represents the step of the diffusion model, distinct from the input video time index $t$.

However, naively using this priors can result in artifacts. While the diffusion priors offers information for unseen views, its accuracy and consistency are not guaranteed across all time. 
Although the Gaussian Splatting process largely preserves temporal consistency due to fixed color and opacity, incorrect transformations may still occur.
To address this issue, drawing inspiration from traditional methods~\cite{rigidloss_bouaziz2023projective} in physical simulations, we propose the rigid regularization $\mathcal{L}_\text{rig}$ to constrain the transformation:
\begin{align}
    \mathcal{L}_\text{rig} = ||J^t-UV^T||_1
\end{align}
Where $U$ and $V$ are the unitary matrices resulting from the Singular Value Decomposition (SVD) of $J^t$.

$UV^T$  can be considered the most similar rotation matrix to $J^t$, and this rigid regularization aims to approximate the transformation as closely as possible to a rigid transformation.
We also discover that this regularization can mitigate the adverse effects caused by inconsistencies, which arise from imperfect foreground segmentation.

\subsection{Training objective}
Besides $\mathcal{L_\text{sds}}$ and $\mathcal{L}_\text{rig}$, we also employ reconstruction loss $\mathcal{L_\text{recon}}$ to compare the input with the rendered results using a combination the L1 loss $\mathcal{L}_1$, the mask loss $\mathcal{L}_{mask}$ and the perceptual loss~\cite{LPIPS} $\mathcal{L}_{\text{vgg}}$.

In summary, the final loss can be written:
\begin{align}
    \mathcal{L} = \lambda_1\mathcal{L_\text{sds}}+ \lambda_2\mathcal{L}_\text{rig} + \lambda_3\mathcal{L}_{\text{vgg}} + \lambda_4\mathcal{L}_1 + \lambda_5\mathcal{L}_{mask}
\end{align}
where $\lambda$ is the weight of each loss.

\begin{figure}[h!]
  \centering
  \includegraphics[width=0.95\linewidth]{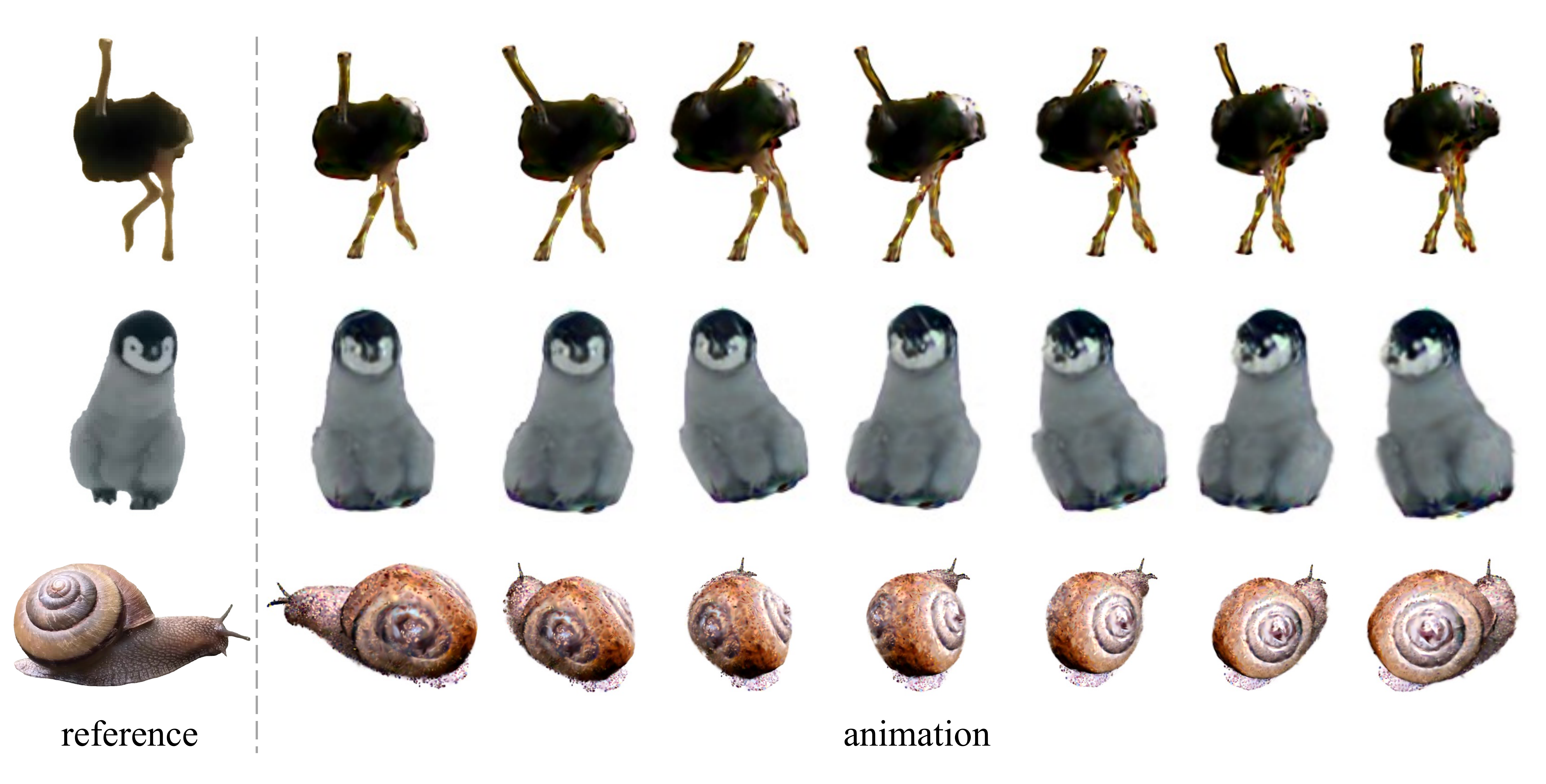}
  \caption{
  Animation results. Our method supports manual manipulation for generating animation.
  }
  \label{fig:animation}
\end{figure}

\begin{figure}[h!]
  \centering
  \includegraphics[width=0.95\linewidth]{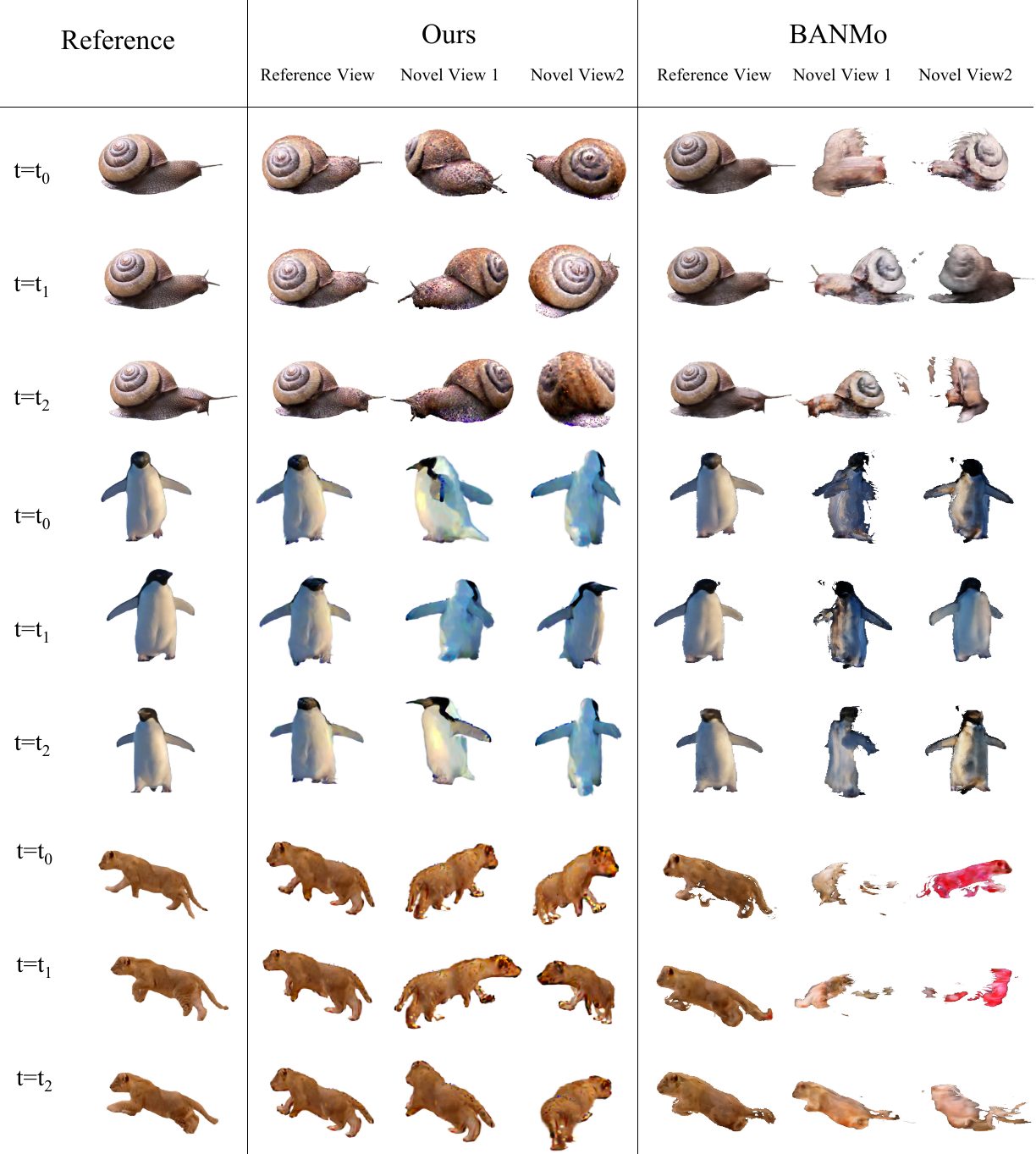}
  \caption{
  Qualitative results.
  Compared with BANMo~\cite{Banmo_yang2022banmo}, our method demonstrates better performance, exhibiting enhanced geometric detail and texture richness. BANMo fails to produce reasonable 3D shapes, instead getting merely a 2D plane that overfits the input video.
  }
  \label{fig:supp_comp_new}
\end{figure}

\section{Experiment}
\subsection{Implementation Details}
During the experiment, it was discovered that training all Gaussians and the neural network-based bones simultaneously resulted in converging toward a noisy canonical space. Consequently, a random frame of the input video was selected as a reference for the preliminary warm-up training of the canonical space. 
After warming up, we start the joint training of all the frames.
We employ operations similar to those described by \cite{3dgs_kerbl20233d}—specifically, split, clone, and prune—to manipulate the Gaussians and enhance rendering quality. These controls are only applied during the warm-up stage.

During the Training, we find out that training all frames together immediately after warming up may cause mistakes. When a target frame is too far from the chosen frame for warming up, the losses may drive the bones to a completely wrong position, since there's a huge difference between the target pose and the current pose. On considering this problem, we adopt a strategy to limit the training frames: At first, only the chosen frame and the other 6 frames nearby are added to the training set. During the training progress, we gradually add all frames to the joint training, making full use of the temporal continuity of motions between adjacent frames.

Throughout the experiments, the weights of our loss terms $\lambda_1$, $\lambda_2$, $\lambda_3$, $\lambda_4$ and $\lambda_5$ are $10^{-4}$, $10^{-1}$, $10^{-1}$, $10^{-1}$ and $1$, respectively.
For the SDS loss, during the warm-up stage, we linearly decrease the $\tau $ from $0.98$ to $0.02$. 
In the joint training stage, 
the warm-up phase ensures a good initialization, allowing $\tau$ to be linearly reduced from $0.5$ to $0.02$.
We run all experiments on a single 40 GB A100 GPU.
Further details are available in the supplementary material.
\subsection{Dataset}
To evaluate our method both qualitatively and quantitatively, we collect 40 videos from the Internet and DAVIS~\cite{davis_Perazzi2016} featuring a variety of animal species exhibiting natural behaviors. To specifically assess our method's capability in reconstructing unseen parts of animals, we opt for videos with lower view coverage. For foreground segmentation, we utilize SAM-Track~\cite{seg_and_track_cheng2023segment}, which is effective across a broad range of object categories. 

\subsection{Quantitative Results}
We select BANMo~\cite{Banmo_yang2022banmo}, which closely matches our setup and is open-source, as the baseline. 
Due to the absence of ground truth for novel view and keypoint information, we utilize CLIP~\cite{clip_radford2021learning} for evaluation.
Initially, for each frame of the input video, the input image served as the reference image, and we generate a novel view image at this time index.
Subsequently, we compute the CLIP image embedding for both the novel view images and the reference images, employing their cosine similarity as the evaluation metric. Consistency was further assess by calculating the cosine similarity between pairs of novel view images.
Additionally, we compare both the training time and rendering time between our method and the baseline. 

The results are presented in Table~\ref{tab:compare_banmo}. 
Our method achieves higher similarity to the reference images and better consistency in novel view images. 
Additionally, it significantly reduced the optimization time from hours to minutes, enabling real-time rendering.
It is important to note that while BANMo training requires 2 GPUs, our method operates with just a single GPU.
\setlength{\tabcolsep}{6pt}
\begin{table}[t]
\renewcommand\arraystretch{1.3} 
\begin{center}
\caption{
Quantitative results with BANMo~\cite{Banmo_yang2022banmo}.
Our method demonstrates superior similarity to the reference images and enhanced consistency in novel view images. Furthermore, it substantially decreases optimization time from hours to minutes and facilitates real-time rendering. Notably, while BANMo training necessitates the use of 2 GPUs, our method operates on a single GPU.
}
\label{tab:compare_banmo}
\begin{tabular}{c|cccc}
\hline
 & CLIP-reference & CLIP-novel &Training  &  rendering  \\
\hline
BANMo& 0.8337 &  0.8795& 12h (2 GPUs)& 0.1 FPS\\ 
\hline
Ours & 0.9023 & 0.9292  & 40min (1 GPU)& 61 FPS\\ 
\hline
\end{tabular}

\end{center}
\end{table}

\begin{figure}[h!]
  \centering
  \includegraphics[width=0.95\linewidth]{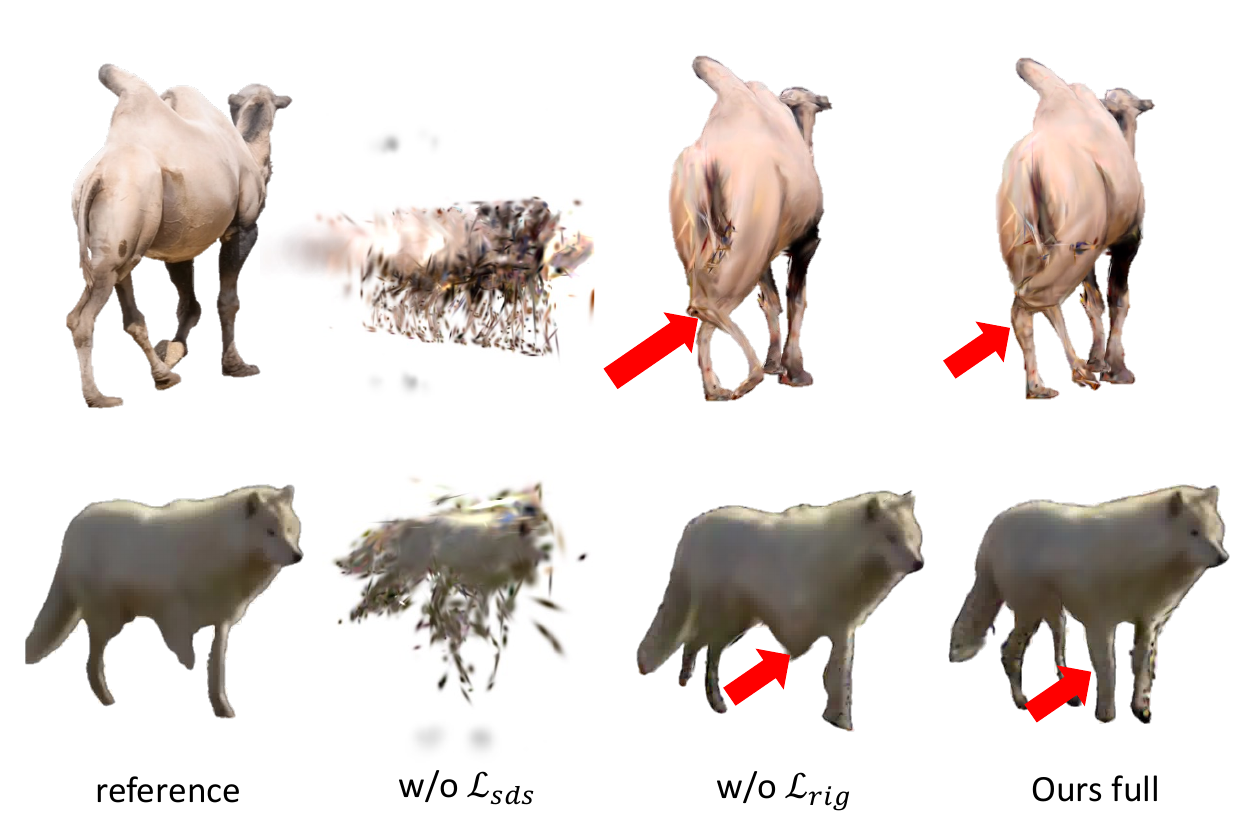}
  \caption{
  Ablation study.
  Without diffusion priors, the model cannot obtain reasonable results. In the absence of rigid regularization, the transformation results in artifacts, such as a tortuous leg. Even when the reference image has an incorrect foreground segmentation, our method, employing rigid regularization, still manages to achieve reasonable outcomes.
  }
  \label{fig:ablation}
\end{figure}

\begin{figure}[h!]
  \centering
  \includegraphics[width=0.95\linewidth]{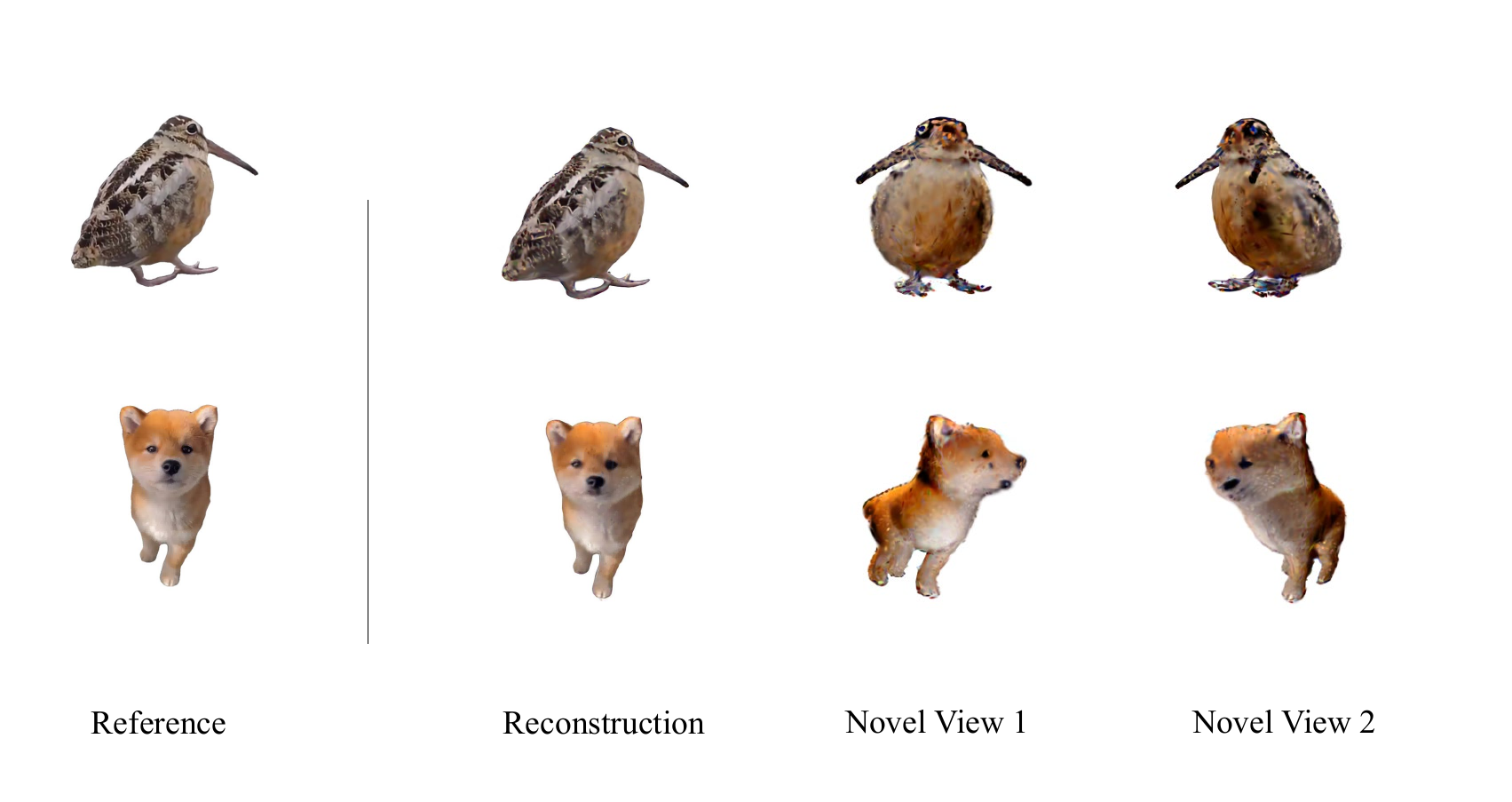}
\caption{Failure case. Despite the diffusion model's advanced capabilities beyond category-level classification, it may still generate artifacts; although the rigid regularization can sometimes mitigate these issues. We plan to address these issues in future research.}
  \label{fig:failure_case}
\end{figure}

\subsection{Qualitative Results}
We present qualitative results in Figure~\ref{fig:compare} and Figure~\ref{fig:supp_comp_new}, illustrating the outcomes rendered across different time steps and camera views. Our approach is compared with BANMo, demonstrating that our method achieves enhanced detail in both geometry and texture. Furthermore, our approach demonstrates superior performance in novel view synthesis, where BANMo struggles to produce accurate 3D shapes, often resulting in the reconstruction of merely a 2D plane that overfits the input video. 
Additionally, animation results showcased in Figure~\ref{fig:animation} illustrate the results of manual manipulation of the 3D models.
More results can be found in the supplementary material.
\subsection{Ablation Study}
The primary contribution of our paper is the utilization of diffusion priors combined with rigid regularization to enhance the application of diffusion priors. As illustrated in Figure~\ref{fig:ablation}, without diffusion priors, the model can not obtain reasonable results. In the absence of rigid regularization, the transformation results in artifacts, such as a tortuous leg. 
Even when the reference image possesses an incorrect foreground segmentation, our method, with rigid regularization, still manages to achieve reasonable results.


\section{Limitation}
Although our method can achieve notably positive results, it is not without limitations. Firstly, while our method utilizes diffusion to compensate for unseen view information, the motion information is still dependent on the input video. Therefore, if the objects in the video do not exhibit sufficient motion, our method may not accurately learn the animatable ability. 
The motion diffusion model can potentially be integrated into the pipeline to address this issue.
Secondly, the generation of unseen views relies on a diffusion model. Although this model has capabilities beyond the category level, it sometimes produces artifacts, as shown in Figure~\ref{fig:failure_case}. While the rigid loss can occasionally mitigate these issues, artifacts remain a concern. 
We aim to address them in future work.
\section{Conclusion}
In conclusion, this study introduces BAGS, an innovative approach for creating animatable 3D models from monocular videos through Gaussian Splatting with diffusion priors, marking an advancement in 3D reconstruction technology. Unlike previous methods that depend heavily on extensive view coverage and incur high computational costs, our method significantly enhances efficiency in training and rendering processes. The integration of diffusion priors enables the learning of 3D models from limited viewpoints, while rigid regularization further optimizes the utilization of these priors. Comparative evaluations with various real-world videos underscore our method's superior performance against existing state-of-the-art techniques.

\bibliographystyle{splncs04}

\bibliography{main}
\end{document}